\documentclass[conference]{IEEEtran}
\IEEEoverridecommandlockouts
\usepackage{cite}
\usepackage{amsmath,amssymb,amsfonts}
\usepackage{algorithmic}
\usepackage{graphicx}
\usepackage{textcomp}
\usepackage{xcolor}
\usepackage{multirow} 
\usepackage{tikz} 
\usepackage{fancybox,framed}
\usepackage{tcolorbox}
\def\BibTeX{{\rm B\kern-.05em{\sc i\kern-.025em b}\kern-.08em
    T\kern-.1667em\lower.7ex\hbox{E}\kern-.125emX}}
    
\newcommand{\mybox}[4]{
    \begin{figure}[h]
        \centering
    \begin{tikzpicture}
        \node[anchor=text,text width=\columnwidth-0.5cm, draw, rounded corners, line width=0.5pt, fill=#3, inner sep=2mm] (big) {\\#4};
        \node[draw, rounded corners, line width=.5pt, fill=#2, anchor=west, xshift=4mm] (small) at (big.north west) {#1};
    \end{tikzpicture}
    \end{figure}
}

\begin{document}

\IEEEoverridecommandlockouts
\IEEEpubid{\begin{minipage}[t]{\textwidth}\ \\[1pt]
        \centering\footnotesize{\begin{tcolorbox}[left = 0.5mm, right = 0.5mm, top = 0.5mm, bottom = 0.5mm]\copyright This work has been submitted to the IEEE for possible publication. Copyright may be transferred without notice, after which this version may no longer be accessible.\end{tcolorbox}}
\end{minipage}}

\title{A Theoretical Analysis of Analogy-Based Evolutionary Transfer Optimization\\
}

\author{\IEEEauthorblockN{Xiaoming Xue\IEEEauthorrefmark{1},
Liang Feng\IEEEauthorrefmark{2}, Yinglan Feng\IEEEauthorrefmark{1}, Rui Liu\IEEEauthorrefmark{1}, Kai Zhang\IEEEauthorrefmark{3} and
Kay Chen Tan\IEEEauthorrefmark{1}}
\IEEEauthorblockA{Department of Data Science and Artificial Intelligence, The Hong Kong Polytechnic University, Hong Kong SAR\IEEEauthorrefmark{1}\\
College of Computer Science, Chongqing University, China\IEEEauthorrefmark{2}\\
School of Civil Engineering, Qingdao University of Technology, China\IEEEauthorrefmark{3}\\
Emails: \{xiaoming.xue, yinglfeng, rui-ricky.liu, kctan\}@polyu.edu.hk, liangf@cqu.edu.cn, zhangkai@qut.edu.cn}}


\maketitle


\begin{abstract}

Evolutionary transfer optimization (ETO) has been gaining popularity in research over the years due to its outstanding knowledge transfer ability to address various challenges in optimization.
However, a pressing issue in this field is that the invention of new ETO algorithms has far outpaced the development of fundamental theories needed to clearly understand the key factors contributing to the success of these algorithms for effective generalization.
In response to this challenge, this study aims to establish theoretical foundations for analogy-based ETO, specifically to support various algorithms that frequently reference a key concept known as \emph{similarity}.
First, we introduce analogical reasoning and link its subprocesses to three key issues in ETO.
Then, we develop theories for analogy-based knowledge transfer, rooted in the principles that underlie the subprocesses.
Afterwards, we present two theorems related to the performance gain of analogy-based knowledge transfer, namely unconditionally nonnegative performance gain and conditionally positive performance gain, to theoretically demonstrate the effectiveness of various analogy-based ETO methods.
Last but not least, we offer a novel insight into analogy-based ETO that interprets its conditional superiority over traditional evolutionary optimization through the lens of the no free lunch theorem for optimization.

\end{abstract}



\section{Introduction}

Recent years have witnessed a surge of research interest in enhancing the performance of evolutionary algorithms through knowledge transfer, a domain referred to as evolutionary transfer optimization (ETO)~\cite{tan2021evolutionary}.
Based on the conceptual realizations outlined in~\cite{gupta2017insights}, ETO can be broadly categorized into three groups: sequential transfer~\cite{feng2017autoencoding,xue2023solution}, multitasking~\cite{gupta2015multifactorial,feng2018evolutionary}, and multiform optimization~\cite{feng2024review}, all of which have been widely used to address various challenges in optimization, including dynamic environment~\cite{jiang2017transfer}, high-dimensional variables~\cite{liu2022evolutionary}, conflicting objectives~\cite{li2024multiobjective}, multi-modal landscape~\cite{gao2023distributed}, etc.

The key feature of ETO that distinguishes it from traditional evolutionary optimization is \emph{knowledge transfer}, which aims to transfer valuable information across different tasks to enhance search performance.
Despite the clear objective, it is important to note that knowledge transfer itself is principle-agnostic---one cannot grasp the fundamental rationale behind its success by examining its literal meaning alone.
The primary reason is that knowledge transfer emphasizes the process of transferring knowledge rather than elucidating why such transfer is effective.
Indeed, various reasoning methods have been widely employed to rationalize knowledge transfer in ETO, including analogy~\cite{kawakami2024evolutionary,ji2025similar}, generalization~\cite{liaw2017evolutionary}, abduction~\cite{feng2023multi}, and prediction~\cite{friess2021predicting,wu2024learning,wang2025learning}, among which analogical reasoning has been gaining remarkable popularity due to its central concept---\emph{similarity}---which is frequently used to inform decisions related to knowledge transfer.
Related ETO works surrounding this concept include benchmark design~\cite{xue2023scalable,hou2024bridging}, metric analysis~\cite{gupta2016landscape}, source selection~\cite{lin2024multiobjective}, adaptive knowledge transfer~\cite{cai2021evolutionary}, task adaptation~\cite{bali2017linearized}, among others.

Despite the significant advances in algorithm design for ETO, the development of its fundamental theories has lagged considerably behind~\cite{scott2023first}.
This gap presents challenges for practitioners in understanding the essential ingredients required for the success of various algorithms, ultimately hindering their performance when generalizing to new scenarios.
In response to this situation, Scott and De Jong~\cite{scott2023first} challenge the plausibility of universally effective knowledge transfer by proving its compliance with the no free lunch theorem for optimization~\cite{wolpert1997no}.
In addition, the theoretical effectiveness of adaptive knowledge transfer has been scatteredly explored in various ETO algorithms that transfer particular forms of knowledge, including the optimized solution~\cite{bali2019multifactorial,xue2024surrogate}, evaluated solution~\cite{liu2024extremo}, surrogate model~\cite{min2017multiproblem}, and probabilistic distribution model~\cite{da2018curbing}, among others.
More recently, the theoretical validity of big-source retrieval for ensuring positive transfer has been reported in~\cite{cao2024competitive}.
However, all these exploratory advances in theory were made in parallel and have not been unified under a common set of fundamentals, perpetuating the gap between theory and practice in ETO.

To address the identified research gap, this study aims to establish the theoretical foundations for analogy-based ETO, specifically to support various ETO algorithms that utilize analogical reasoning for knowledge transfer.
First, we define analogical reasoning within the context of ETO and link its three subprocesses to three key issues in ETO: 1) retrieval for determining what to transfer; 2) mapping for guiding how to transfer; and 3) evaluation for deciding when to transfer.
Next, we establish the theoretical foundations for analogy-based knowledge transfer grounded in the principles underlying the three subprocesses.
Afterwards, we present two important theorems concerning the performance gain achieved through analogy-based knowledge transfer: unconditionally nonnegative performance gain and conditionally positive performance gain.
Lastly, we provide a novel insight into ETO that not only demonstrates its compliance with the no free lunch theorem for optimization but also fundamentally interprets the conditional superiority of ETO over traditional evolutionary optimization.

The rest of this paper is structured as follows.
Section \ref{sec:pre} provides a brief introduction to the definitions of ETO and analogical reasoning.
Section \ref{sec:akt} outlines the theoretical foundations of analogy-based knowledge transfer within ETO.
In Section \ref{sec:pg}, the two theorems are presented alongside our novel insights into ETO from the perspective of no free lunch.
Finally, Section \ref{sec:con} concludes this paper.


\section{Preliminaries}
\label{sec:pre}

In this section, we first introduce the definition of ETO and outline its three key issues.
Then, we introduce analogical reasoning and link its subprocesses to the key issues of ETO.

\subsection{Evolutionary Transfer Optimization}

ETO is an optimization paradigm that empowers evolutionary optimization to transfer knowledge across tasks toward enhanced search performance~\cite{tan2021evolutionary}, which typically involves addressing three key issues of knowledge transfer~\cite{scott2024varying,villar2025transfer}:

\begin{itemize}
\item \emph{What to Transfer}: It concerns identifying the appropriate knowledge to be transferred across optimization tasks to improve search performance effectively.

\item \emph{How to Transfer}: It involves adopting an effective method for transferring knowledge to maximize its benefit for the intended recipient (the target task).

\item \emph{When to Transfer}: It involves ascertaining the timing for knowledge transfer to maximize positive outcomes while minimizing the risk of negative transfer.
\end{itemize}

Various methods of reasoning can address the three issues, with analogical reasoning gaining the most research attention.
This popularity primarily stems from its inherent applicability to source-target instances that share certain similarities.

\subsection{Analogical Reasoning}

Analogical reasoning is a cognitive process that involves recognizing structural similarities between two subjects and making inferences based on these parallels~\cite{gentner2012analogical}.
This method of reasoning allows individuals to transfer knowledge from one context (referred to as the source) to another (the target), facilitating deeper understanding and efficient problem-solving.
Its fundamental logic for rationalizing knowledge transfer in ETO can be articulated as follows:

\mybox{Rationale}{black!20}{black!5}{If two tasks $T_a$ and $T_b$ are similar in terms of $w$, and $T_a$ has been learned to possess knowledge $v$ related to $w$, then $T_b$ is likely to benefit from $v$.}

\begin{figure}[htbp]
	\centerline{\includegraphics[width=2.2in]{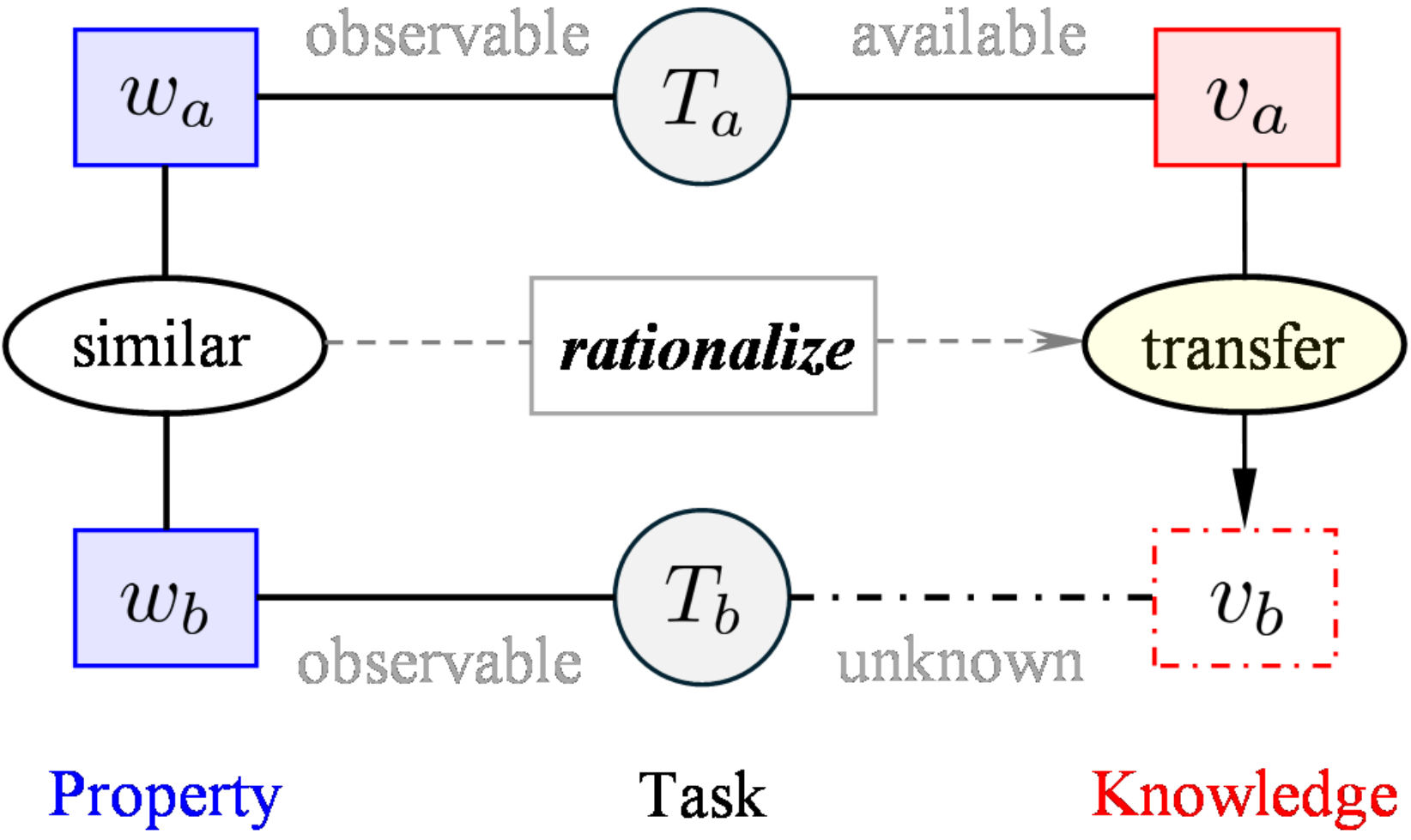}}
	\caption{Analogy-based rationale for knowledge transfer.}
	\label{fig:analogy}
\end{figure}

This rationale underpins many ETO algorithms that transfer knowledge $v$ from one task to another based on their similarity concerning property $w$, as illustrated in Fig. \ref{fig:analogy}.
Typically, the property $w$ is commonly observable (e.g., population data), while the knowledge $v$ is available for the source task but remains unknown or insufficient for the target task (e.g., the optimal solution).
This situation allows for enhanced problem-solving in the target context through proper knowledge transfer from the source.
With assured association between $w$ and $v$, the rationale's effectiveness hinges on the following three subprocesses of analogical reasoning~\cite{gentner2012analogical}:
\begin{itemize}
\item \emph{Retrieval}: When multiple options $\boldsymbol{v}=\lbrace v_i\mid1\le i\le k\rbrace$ are available for transfer, the knowledge that provides the maximum benefit for the target task should be retrieved.

\item \emph{Mapping}: For a knowledge $v$ ready for transfer, the target task can benefit more if it is properly adapted to $v'$.

\item \emph{Evaluation}: For a knowledge $v$ ready to take effect on the target task, the knowledge transfer should be permitted only if the target task can indeed benefit from $v$.
\end{itemize}

It is evident that these subprocesses align closely with the three key issues in ETO: retrieval for what to transfer, mapping for how to transfer, and evaluation for when to transfer.
Next, we will analyze these subprocesses by examining them as theoretical solutions to their respective key issues.

\section{Analogy-Based Knowledge Transfer}

\label{sec:akt}

In this section, we will establish the theoretical foundations for analogy-based knowledge transfer in ETO by separately examining the principles underlying the three subprocesses.
Besides, we will show that the validity of these principles can be theoretically assured through clear and definitive premises.

\subsection{What to Transfer---Retrieval}

The retrieval subprocess is responsible for selecting or weighting promising knowledge among multiple candidates based on their observable similarities to the target task, the principle behind which can be explicitly stated as follows:

\mybox{Principle 1}{black!20}{black!5}{The greater the similarity between two tasks $T_a$ and $T_b$, the more useful $v_a$ from $T_a$ will be for the target task $T_b$.}

The validity of Principle 1 can be firmly supported by two properties of the relationship between the measured similarity $s\left(w\right)\in S$ and the knowledge usefulness $u\left(v\right)\in U$:

\begin{itemize}

\item \emph{Bijectivity}: There is a one-to-one correspondence between the elements in $S$ and $U$, i.e., $S\leftrightarrow U$, indicating their association and enabling the deterministic inference of every element $u\in U$ from a unique element $s\in S$.

\item \emph{Continuity}: A small variation of the measured similarity $s$ always induces a small variation of the knowledge usefulness $u$ to be inferred, i.e., $\lim_{s\to s\left(w_a\right)}u=u\left(v_a\right)$, indicating the reliability of using $s$ to infer $u$.

\end{itemize}

These two properties, serving as premises, are essential to rationalize Principle 1, jointly leading to a monotone function between $s$ and $u$.
It allows us to infer $u$ based on $s$:

\mybox{Theorem 1}{yellow!20}{yellow!5}{With the bijectivity and continuity, there must exist a monotonic function $f:S\to U$ such that the rank of any $u_*\in U$ can be deterministically inferred according to $s_*$.}

\begin{IEEEproof}[Proof] First, we assume that $S$ and $U$ are two ordered sets, without loss of generality.
Next, we will use proof by contradiction to demonstrate that any bijective and continuous mapping $f$ between $S$ and $U$ must be monotonic.

Suppose, for contradiction, that $f$ is not monotonic.
This means there exist points $s_1, s_2, s_3\in S$ such that $s_1<s_2<s_3$ but $\left(u_2-u_1\right)\left(u_3-u_2\right)<0$, which means $u_1, u_3<u_2$ or $u_2<u_1, u_3$.
Since $f$ is continuous, by the intermediate value theorem, these exist some $u_i\in\left(\max\lbrace u_1,u_3\rbrace, u_2\right)$ (or $u_i\in\left( u_2, \min\lbrace u_1,u_3\rbrace\right)$) such that $f\left(u_i\right)$ takes on both $\left(s_1,s_2\right)$ and $\left(s_2,s_3\right)$, which contradicts the bijectivity of $f$.
This contradiction shows that $\left(u_2-u_1\right)\left(u_3-u_2\right)\ge0,\,\forall s_1<s_2<s_3\in S$.
Additionally, the bijectivity of $f$ shows that $u_1\ne u_2\ne u_3,\,\forall s_1\ne s_2\ne s_3\in S$, thereby leading to
\begin{equation}
\left(u_2-u_1\right)\left(u_3-u_2\right)>0,\,\forall s_1<s_2<s_3\in S.
\label{eq:mono}
\end{equation}

With Eq. \eqref{eq:mono}, one can conclude that $f: S\to U$ is either strictly increasing or decreasing.
Throughout this paper, we consider $f$ to be strictly increasing to align with the statement in Principle 1, without loss of generality.
In the case of a decreasing $f$, it can be transformed into an increasing function by applying the negative operation, i.e., $-f$.
Next, the rank of any element $s\in S$ is defined as $r\left(s\right)=\lbrace\mathrm{card}\left(S^-\right)| s^-\in S:s^-<s\rbrace$, where $\mathrm{card}\left(S^-\right)$ denotes the cardinal number of $S^-$.
Suppose $f$ is strictly increasing, we have
\begin{equation}
\begin{cases}
f\left(s\right)>f\left(s^-\right)\mid\forall s^-\in S:s^-<s,\\
f\left(s\right)<f\left(s^+\right)\mid\forall s^+\in S:s<s^+.
\end{cases}
\label{eq:cardinality}
\end{equation}

The rank of $u$ associated with $s$ is simply
\begin{equation}
\small
\begin{split}
r(u)&=\lbrace\mathrm{card}\left(U^-\right)\mid u^-=f\left(s^-\right)\in U:f\left(s^-\right)<f\left(s\right)\rbrace\\
&=\lbrace\mathrm{card}\left(S^-\right)| s^-\in S:s^-<s\rbrace\\
&=r\left(s\right).
\end{split}
\label{eq:rank}
\end{equation}


\end{IEEEproof}

\begin{figure}[htbp]
	\centerline{\includegraphics[width=3.4in]{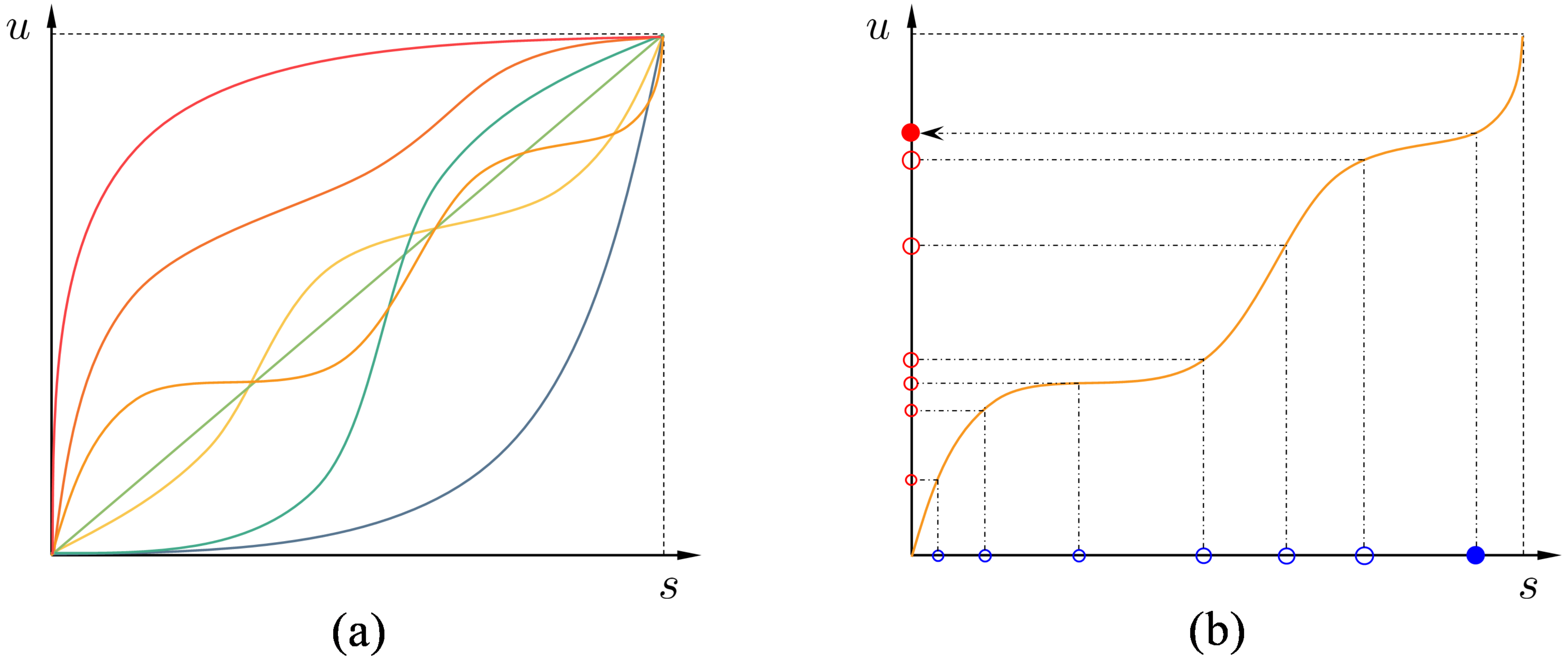}}
	\caption{Illustrations of some monotonic functions $f:S\to U$ and the retrieval process in Eq. \eqref{eq:retrieval}: (a) the monotonic functions; (b) the retrieval process.}
	\label{fig:retrieval}
\end{figure}

Theorem 1 allows for the accurate identification of the most promising knowledge among multiple options through similarity qualification, as formulated by
\begin{equation}
v_p=\underset{v\in\boldsymbol{v}}{\mathrm{arg\,max\,\,}}u\left(v\right)=\underset{v;w\in\boldsymbol{v};\boldsymbol{w}}{\mathrm{arg\,max\,\,}}s\left(w\right),
\label{eq:retrieval}
\end{equation}
where $v_p$ is the identified promising knowledge.
The identification process in Eq. \eqref{eq:retrieval} is quite general, encompassing both hard selection~\cite{chen2019adaptive,cao2024global}---which identifies the knowledge with the highest similarity---and soft selection~\cite{da2018curbing,bali2019multifactorial,liu2024extremo}, which retrieves knowledge through a mixture model that weights the varying contributions of different candidates\footnote{We choose to interpret the weights as contributions rather than similarities, since knowledge retrieval and transfer occur at the level of mixture model, where Eq. \eqref{eq:retrieval} is typically fulfilled through likelihood maximization.}.

To demonstrate, we present several examples of monotonic functions $f:S\to U$ in Fig. \ref{fig:retrieval}(a), along with an illustration of the retrieval process in Fig. \ref{fig:retrieval}(b), wherein the measured similarities and knowledge usefulness are represented by blue and red circles, respectively, with their sizes gradually increasing to indicate their ranks.
It is evident that the most useful knowledge, represented by the red solid circle, can be consistently identified based on the highest measured similarity, as indicated by the blue solid circle.

\subsection{How to Transfer---Mapping}
\label{akt:how}
Informed by the observable similarity, the mapping subprocess is responsible for determining an appropriate way of transferring knowledge to maximize its usefulness for the target task, the principle behind which is given by

\mybox{Principle 2}{black!20}{black!5}{The usefulness of knowledge $v_a$ from $T_a$ for the target task $T_b$ can be enhanced by making $T_a$ more similar to $T_b$ through appropriate adaptation.}

The validity of Principle 2 can be firmly supported by two preconditions: realizability and optimality, as outlined below.

\begin{figure}[htbp]
	\centerline{\includegraphics[width=3.4in]{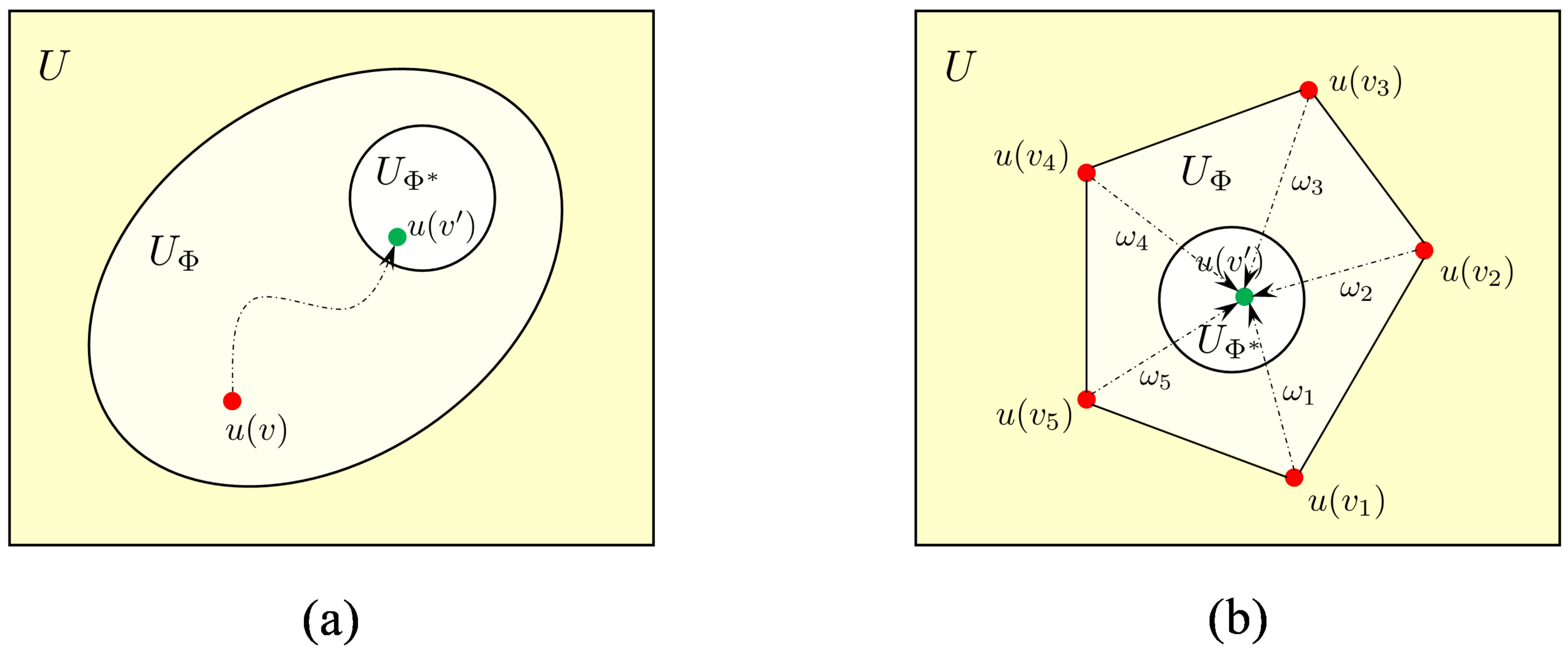}}
	\caption{An illustrative comparison between explicit mapping learning and mixture-based implicit mapping: (a) the explicit mapping; (b) the mixture-based mapping.}
	\label{fig:mm}
\end{figure}

\begin{itemize}

\item \emph{Realizability}: There exists a hypothesis set of mappings\footnote{These mappings should preserve the association between $w$ and $v$---Each adapted task remains a valid member of the task set, i.e., $\phi\left(T\right)\in\mathbb{T},\forall\phi\in\Phi$.} $\Phi=\lbrace\phi:\mathbb{T}\to\mathbb{T}\rbrace$ available to adapt $T:\lbrace v,w\rbrace$ to $T':\lbrace v',w'\rbrace$ such that $S^*=\lbrace s\mid s\left(w'\right)>s\left(w\right):w'=\phi\left(w\right),\forall \phi\in\Phi\rbrace$ is not empty, indicating the realizability of higher observable similarity through appropriate mapping operations.
The set of mappings associated with the non-empty set is denoted by $\Phi^*\subseteq\Phi$.

\item \emph{Optimality}: There exists an algorithm that consistently returns a candidate mapping $\phi^*$ such that it belongs to $\Phi^*$, i.e., $\phi^*\in\Phi^*$, ensuring that the observable similarity is effectively enhanced after the mapping.

\end{itemize}

These two preconditions are essential for justifying Principle 2, collectively leading to an important theorem regarding the enhancement of knowledge usefulness through improved observable similarity, as formally stated by

\mybox{Theorem 2}{yellow!20}{yellow!5}{With the realizability and optimality for a hypothesis set of mappings $\Phi$, the usefulness of knowledge $v$ can be enhanced by adapting the source task using $\phi\in\Phi$ to achieve improved observable similarity.}

\begin{IEEEproof}[Proof] With the realizability and Theorem 1, we have

\begin{equation}
u\left(\phi\left(v\right)\right)>u\left(v\right), \forall \phi\in \Phi^*.
\end{equation}

With the optimality, one can always obtain $\phi^*\in\Phi^*$ such that $u\left(\phi^*\left(v\right)\right)>u\left(v\right)$.
This indicates that the usefulness of knowledge $v$, adapted by $\phi^*$, can be effectively improved.

\end{IEEEproof}

Theorem 2 lays the foundation for maximizing the benefit of the knowledge at hand by properly adapting it to enhance the observable similarity to the target task, as formulated by
\begin{equation}
v_m=\underset{\phi\left(v\right)\in\Phi\left(v\right)}{\mathrm{arg\,max\,\,}}u\left(\phi\left(v\right)\right)=\underset{\phi\left(v;w\right)\in\Phi\left(v;w\right)}{\mathrm{arg\,max\,\,}}s\left(\phi\left(w\right)\right),
\label{eq:mapping}
\end{equation}
where $v_m$ denotes the adapted form of $v$.
The mapping process in Eq. \eqref{eq:mapping} is fairly general, including both explicit mapping learning~\cite{bali2017linearized,feng2018evolutionary,xue2020affine}---which establishes an explicit source-to-target mapping based on specific similarity enhancement criteria---and implicit mapping based on mixture model~\cite{min2017multiproblem,friess2020improving}, which induces an implicit mapping through the optimal convex combination of individual knowledge candidates.

\begin{figure}[htbp]
	\centerline{\includegraphics[width=2.3in]{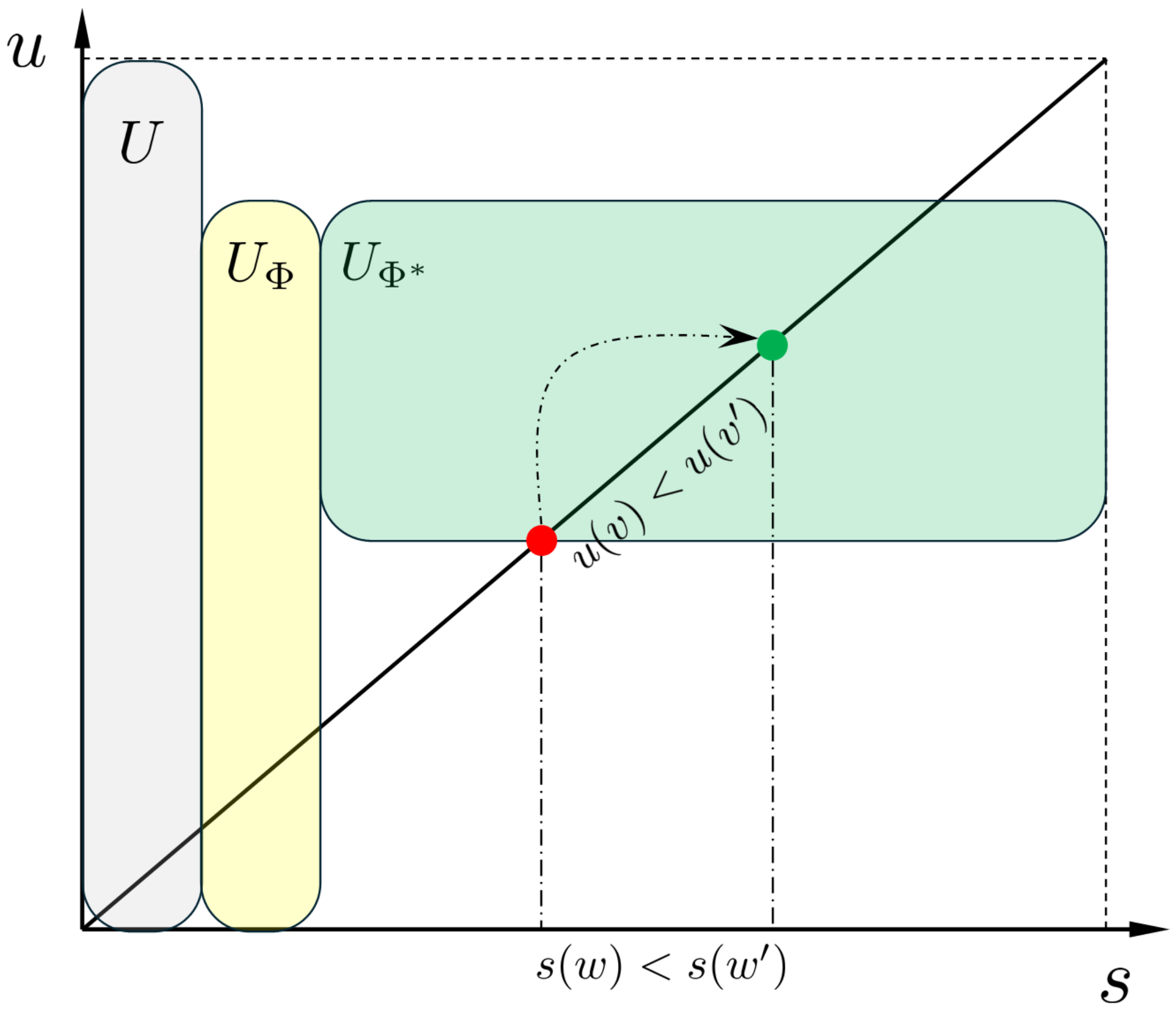}}
	\caption{An illustration of the mapping process in Eq. \eqref{eq:mapping}.}
	\label{fig:mapping}
\end{figure}

To illustrate, we compare the explicit mapping learning and mixture-based implicit mapping in Fig. \ref{fig:mm}, where $U_\Phi=\lbrace u\mid u\left(\phi\left(v\right)\right): \forall\phi\in\Phi\rbrace$ denotes the realizable usefulness of knowledge through $\Phi$, $U_{\Phi^*}=\lbrace u\mid u\left(\phi\left(v\right)\right)>u\left(v\right): \forall\phi\in\Phi\rbrace$ denotes the improved usefulness of knowledge through $\Phi^*$.
In the explicit mapping learning shown in Fig. \ref{fig:mm}(a), the realizability depends on the configuration of $\Phi$, while the optimality is determined by the learning algorithm that guides the knowledge $v$ toward $U_{\Phi^*}$.
By contrast, the mixture-based implicit mapping in Fig. \ref{fig:mm}(b) establishes its realizability through the convex hull formed by the mixture of the five individual knowledge candidates.
Its optimality, on the other hand, depends on the learning algorithm that assigns appropriate weights, ensuring that the mixture lies within $U_{\Phi^*}$.

Fig. \ref{fig:mapping} illustrates the mapping process supported by Theorem 2, with readability $U_\Phi$ and optimality $U_{\Phi^*}$ represented by the yellow and green rectangles, respectively.
Without loss of generality, we consider a linear monotonic function $f:S\to U$, as represented by the solid line.
It is evident that enhancing the observable similarity through appropriate mapping can effectively improve the usefulness of knowledge.

\subsection{When to Transfer---Evaluation}

The evaluation subprocess determines whether to transfer knowledge based on its observable similarity to the target task at each moment, thereby providing a complete solution to the issue of `when to transfer' throughout the entire optimization process, whose principle can be explicitly given by
\mybox{Principle 3}{black!20}{black!5}{The usefulness of knowledge $v_a$ from $T_a$ for $T_b$ can be assessed based their observable similarity $s\left(w\right)$.}

The validity of Principle 3 is strengthened as follows. 

\mybox{Theorem 3}{yellow!20}{yellow!5}{There exists a similarity threshold $s^{\mathrm{thre}}$ that can be used to deterministically classify knowledge as useful or useless for the target task---It is considered useful if and only if its similarity to the target task exceeds $s^{\mathrm{thre}}$.}

\begin{IEEEproof}[Proof] Let the current knowledge of the target task at time $\tau$ be denoted by $v^\tau$ and its usefulness represented by $u^\tau\in U$.
Without loss of generality\footnote{One can define the usefulness of knowledge using any other binary states.}, we define knowledge $v$ as useful for the target task if $u\left(v\right)>u^\tau$ and as useless if $u\left(v\right)<u^\tau$.
Based on this notion of usefulness, we can categorize knowledge into two groups:
\begin{equation}
\begin{cases}
V^+=\lbrace v\mid v\in V: u\left(v\right)>u^\tau\rbrace,\\
V^-=\lbrace v\mid v\in V: u\left(v\right)<u^\tau\rbrace,
\end{cases}
\label{eq:useful_useless}
\end{equation}
where $V^+$ and $V^-$ denote the sets of knowledge categorized as useful and useless, respectively.
Since the inverse operation preserves monotonicity, $f^{-1}: U\to S$ must be monotonic.
Then, we can rewrite Eq. \eqref{eq:useful_useless} as follows:

\begin{equation}
\begin{cases}
V^+=\lbrace v\mid v\in V: f^{-1}\left(u\left(v\right)\right)>f^{-1}\left(u^\tau\right)\rbrace,\\
V^-=\lbrace v\mid v\in V: f^{-1}\left(u\left(v\right)\right)<f^{-1}\left(u^\tau\right)\rbrace.
\end{cases}
\label{eq:useful_useless_re}
\end{equation}

By defining a threshold $s^{\mathrm{thre}}$ as $f^{-1}\left(u^\tau\right)$, we have

\begin{equation}
\begin{cases}
V^+=\lbrace v\mid v\in V: s\left(w; v\right)>s^{\mathrm{thre}}\rbrace,\\
V^-=\lbrace v\mid v\in V: s\left(w; v\right)<s^{\mathrm{thre}}\rbrace.
\end{cases}
\label{eq:classification}
\end{equation}

For any knowledge $v\in V$ and $v\ne v^\tau$, it belongs to either $V^+$ or $V^-$ as $V=V^-\cup v^\tau\cup V^+$.
Therefore, this knowledge is considered useful for the target task if and only if its observable similarity to the target task exceeds $s^{\mathrm{thre}}$.

\end{IEEEproof}

Theorem 3 lays the foundation for deciding when to transfer knowledge by assessing its observable similarity to the target task, thereby enabling adaptive transfer that aligns well with its actual usefulness for the target task, as given by

\begin{equation}
v_e=
\begin{cases}
v, &\mathrm{if}\,\,s\left(w;v\right)>s^{\mathrm{thre}},\\
v^\tau, &\mathrm{otherwise},
\end{cases}
\label{eq:evaluation}
\end{equation}
where $v_e$ is the identified knowledge after the evaluation.
It is evident that $v_e=v^\tau$ signifies the cancellation of knowledge transfer.
The evaluation in Eq. \eqref{eq:evaluation} is rather general, encompassing both explicit similarity assessment~\cite{cai2021evolutionary,xue2024surrogate}---which decides when to transfer knowledge based on measured observable similarities---and implicit adaptive transfer based on mixture models~\cite{da2018curbing,bali2019multifactorial,liu2024extremo}, which washes out useless knowledge candidates by applying a mixture weight of zero.

\begin{figure}[htbp]
	\centerline{\includegraphics[width=3.4in]{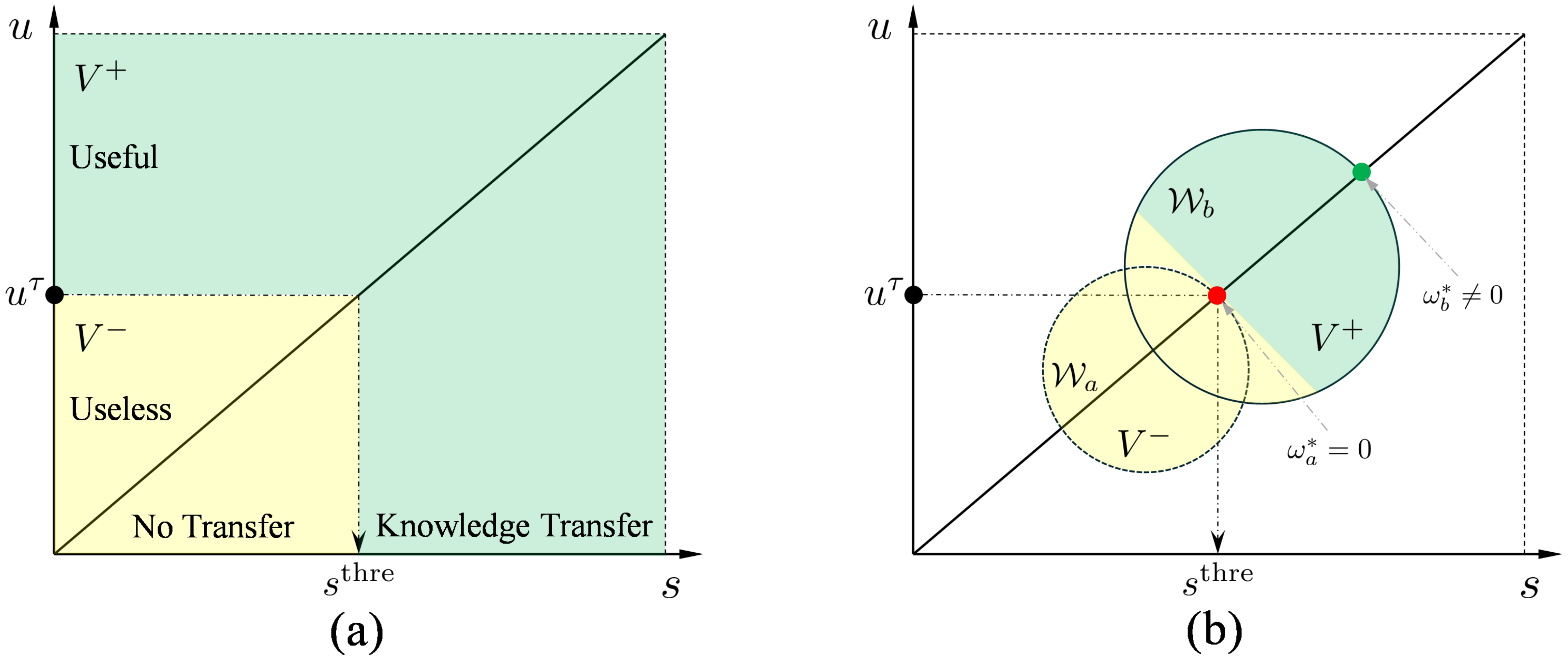}}
	\caption{An illustrative comparison between explicit similarity assessment and mixture-based implicit adaptive transfer: (a) the explicit similarity assessment; (b) the mixture-based adaptive transfer.}
	\label{fig:evaluation}
\end{figure}

An illustrative comparison between the explicit similarity assessment and mixture-based implicit adaptive transfer under Theorem 3 is presented in Fig. \ref{fig:evaluation}, in which the useful and useless knowledge (i.e., $V^+$ and $V^-$) are represented by the yellow and green areas, respectively.
From Fig. \ref{fig:evaluation}(a), one can observe that the similarity threshold indicates when to transfer knowledge.
By comparing the observable similarity with $s^{\mathrm{thre}}$, one can leverage useful knowledge consistently while avoiding the negative impact of useless knowledge.
Unlike the explicit similarity comparison, the mixture transfers knowledge in a more nuanced manner---the contribution of specific knowledge is represented by the assigned weight.
Ideally, by analyzing the entire convex hull, as represented by any of the circles in Fig. \ref{fig:evaluation}(b), the optimal mixture with the highest similarity will be identified, leading to a set of adaptive weights for controlling the contributions of different knowledge.
As shown in Fig. \ref{fig:evaluation}(b), a two-component mixture $v_e=\omega_av_a+\left(1-\omega_a\right) v^\tau$ with the maximum similarity of $s^\mathrm{thre}$ will eliminate the contribution of $v_a$ using $\omega^*_a=0$, as represented by the red solid circle.
A weight of zero in this context signifies the cancellation of knowledge transfer.
In contrast, another mixture $v_e=\omega_bv_b+\left(1-\omega_b\right) v^\tau$ with similarities greater than $s^\mathrm{thre}$ will adopt the optimal mixture with positive $\omega_b$ to achieve the maximum similarity, as indicated by the green solid circle.
A positive weight in this context indicates that knowledge transfer has occurred, contributing a specific value.


\section{Analyses of Performance Gain}
\label{sec:pg}

In this section, we will conduct theoretical analyses of the performance gain achieved through the analogy-based knowledge transfer in Section~\ref{sec:akt}, ultimately leading us to two important theorems, namely \emph{unconditionally nonnegative performance gain} and \emph{conditionally positive performance gain}, as will be separately examined in what follows.
Additionally, we offer a novel insight into ETO that demonstrates its compliance with the no free lunch theorem for optimization~\cite{wolpert1997no}.

\subsection{Unconditionally Nonnegative Performance Gain}

Without loss of generality, we define the performance gain achieved by a knowledge $v$ for the target task as follows:

\begin{equation}
\Delta_v=u\left(v\right)-u^\tau,
\label{eq:pg}
\end{equation}
where $\Delta_v$ denotes the performance gain achieved by $v$ for the target optimization, $u^\tau$ represents the usefulness of the target knowledge till time $\tau$. Now, a pertinent question arises---What protects us from the threat of negative transfer (i.e., $\Delta_v\ge0$)?
A theorem addressing this question is stated as follows:
\mybox{Theorem 4}{yellow!20}{yellow!5}{The performance gain of an analogy-based knowledge transfer method is unconditionally nonnegative if and only if it includes the evaluation subprocess.}

\begin{IEEEproof}[Proof] The three subprocesses described by Eqs. \eqref{eq:retrieval}, \eqref{eq:mapping} and \eqref{eq:evaluation} can be defined as three knowledge-related functions, namely retrieval function $f_\mathrm{r}: V^k\to V$, mapping function $f_\mathrm{m}: V\to V$ and evaluation function $f_\mathrm{e}: V^2\to V$.
Suppose we have $k$ source tasks and their knowledge are given by $\boldsymbol{v}=\lbrace v_i\mid_{1\le i\le k}\rbrace\in V^k$, the greatest lower bounds (i.e., infimums) of the three functions' performance gain are simply
\begin{equation}
\begin{cases}
\alpha^{\inf}_\mathrm{r}=\inf\left(\Delta_{f_\mathrm{r}}\right)=u_{\mathrm{min}}-u^\tau,\\
\alpha^{\inf}_\mathrm{m}=\inf\left(\Delta_{f_\mathrm{m}}\right)=u_{\mathrm{min}}+\delta_u-u^\tau,\\
\alpha^{\inf}_\mathrm{e}=\inf\left(\Delta_{f_\mathrm{e}}\right)=0,
\end{cases}
\label{eq:infs}
\end{equation}
where $\alpha^{\inf}_i$ denotes the infimum of the performance gain of function $i$, $i\in\lbrace\mathrm{r},\mathrm{m},\mathrm{e}\rbrace$, $u_{\mathrm{min}}=\min u\left(f(\boldsymbol{v})\right), \forall\boldsymbol{v}\in V^k$ represents the minimum usefulness in $U$, $\delta_u$ denotes the usefulness improvement achieved by the mapping function $f_\mathrm{m}$.

For any analogy-based knowledge transfer method, we can represent it as a specific composition of $f_\mathrm{r}$, $f_\mathrm{m}$ and $f_\mathrm{e}$.
Specifically, there are 15 compositions in total, including 3 single-subprocess methods $\lbrace f_\mathrm{r}, f_\mathrm{m}, f_\mathrm{e}\rbrace$, 6 double-subprocess methods $\lbrace f_\mathrm{r}\circ f_\mathrm{m}, f_\mathrm{m}\circ f_\mathrm{r}, f_\mathrm{r}\circ f_\mathrm{e}, f_\mathrm{e}\circ f_\mathrm{r}, f_\mathrm{m}\circ f_\mathrm{e}, f_\mathrm{e}\circ f_\mathrm{m}\rbrace$ and 6 triple-subprocess methods $\lbrace f_\mathrm{r}\circ f_\mathrm{m}\circ f_\mathrm{e}, f_\mathrm{r}\circ f_\mathrm{e}\circ f_\mathrm{m}, f_\mathrm{m}\circ f_\mathrm{r}\circ f_\mathrm{e}, f_\mathrm{m}\circ f_\mathrm{e}\circ f_\mathrm{r}, f_\mathrm{e}\circ f_\mathrm{r}\circ f_\mathrm{m}, f_\mathrm{e}\circ f_\mathrm{m}\circ f_\mathrm{r}\rbrace$, as represented by a set of knowledge transfer methods $\mathcal{F}$.
Now, $\forall f\in\mathcal{F}$, we can obtain the infimum of its performance gain as follows:
\begin{equation}
\scriptsize
\alpha^{\inf}_f=
\begin{cases}
\alpha^{\inf}_{\mathrm{r}}, & {\displaystyle\mathrm{if}\,f\in\lbrace f_\mathrm{r}\rbrace},\\
\alpha^{\inf}_{\mathrm{m}}, & {\displaystyle\mathrm{if}\,f\in\lbrace f_\mathrm{m}, f_{\mathrm{r}\circ \mathrm{m}}, f_{\mathrm{m}\circ \mathrm{r}}\rbrace},\\
\alpha^{\inf}_{\mathrm{e}}, & {\displaystyle\mathrm{if}\,f\in\lbrace f_{\textcolor{red}{\mathrm{e}}}, f_{\mathrm{r}\circ \textcolor{red}{\mathrm{e}}}, f_{\textcolor{red}{\mathrm{e}}\circ \mathrm{r}}, f_{\mathrm{m}\circ \textcolor{red}{\mathrm{e}}}, f_{\mathrm{r}\circ \mathrm{m}\circ \textcolor{red}{\mathrm{e}}}, f_{\mathrm{m}\circ \mathrm{r}\circ \textcolor{red}{\mathrm{e}}}, f_{\mathrm{m}\circ \textcolor{red}{\mathrm{e}}\circ \mathrm{r}}\rbrace},\\
\delta_u, & {\displaystyle\mathrm{if}\,f\in\lbrace f_{\textcolor{red}{\mathrm{e}}\circ \mathrm{m}}, f_{\mathrm{r}\circ \textcolor{red}{\mathrm{e}}\circ \mathrm{m}}, f_{\textcolor{red}{\mathrm{e}}\circ \mathrm{r}\circ \mathrm{m}}, f_{\textcolor{red}{\mathrm{e}}\circ \mathrm{m}\circ \mathrm{r}}\rbrace}.
\end{cases}
\label{eq:gains_methods}
\end{equation}

Among the four cases in Eq. \eqref{eq:gains_methods}, $\alpha^{\inf}_{\mathrm{r}}$ and $\alpha^{\inf}_{\mathrm{m}}$ will be negative when $u^\tau>u_{\mathrm{min}}$ and $u^\tau>u_{\mathrm{min}}+\delta_u$, respectively, while $\alpha^{\inf}_{\mathrm{e}}$ and $\delta_u$ remain consistently nonnegative.
By highlighting the evaluation function in red, it becomes clear that a knowledge transfer method $f\in\mathcal{F}$ is safe from negative transfer if and only if the evaluation subprocess is included.
In other words, the evaluation function is both necessary and sufficient for ensuring that an analogy-based knowledge transfer method avoids the risk of negative transfer.

\end{IEEEproof}

The implication of Theorem 4 is quite straightforward: adaptive knowledge transfer, as described in Eq. \eqref{eq:evaluation}, which takes into account the continuously evolving target knowledge throughout the optimization process, is essential for an analogy-based knowledge transfer method to achieve nonnegative performance gain consistently.
This has been theoretically validated across various ETO algorithms with adaptive transfer of different types of knowledge~\cite{da2018curbing,xue2024surrogate,liu2024extremo,min2017multiproblem}.

\subsection{Conditionally Positive Performance Gain}

According to Eq. \eqref{eq:infs}, none of the three subprocesses can consistently achieve a positive performance gain for all cases of $\boldsymbol{v}\in V^k$.
However, if we are willing to invest certain costs or resources, a question arises---What actions should be taken such that positive transfer always occurs (i.e., $\Delta_v>0$)?
A theorem addressing this concern is presented by

\mybox{Theorem 5}{yellow!20}{yellow!5}{The positive performance gain of an analogy-based knowledge transfer method can be conditionally assured with either big-source retrieval or appropriate mapping.}

\begin{IEEEproof}[Proof] Since each of the three subprocesses does not diminish the infimum of knowledge usefulness, the infimum of a composition's performance gain will not be lower than the maximum infimum of its component functions:
\begin{equation}
\small
\alpha^{\inf}_{f_{i\circ j\circ k}}\ge\max\lbrace\alpha^{\inf}_{f_i},\alpha^{\inf}_{f_j},\alpha^{\inf}_{f_k}\rbrace,\,\,\forall i,j,k\in\lbrace \emptyset, \mathrm{r}, \mathrm{m}, \mathrm{e}\rbrace.
\label{eq:inf_pre}
\end{equation}

With Eq. \eqref{eq:inf_pre}, we can streamline our analysis by examining the three functions individually.
Concerning the evaluation function $f_{\mathrm{m}}$, its infimum of performance gain remains zero for all $\boldsymbol{v}\in V^k$.
As for the retrieval function $f_{\mathrm{r}}$, its infimum of performance gain can be effectively lifted as $k\to\infty$ under mild conditions.
Suppose $v$ is a random variable with a positive density $p\left(v\right)$ across the entire domain $V$.
Then, there exists a probability $P^+>0$ corresponding to $V^+$ in Eq. \eqref{eq:useful_useless}:
\begin{equation}
P^+=\int_{v\in V^+}p\left(v\right)\mathrm{d}v.
\label{eq:probability_useful}
\end{equation}

According to the definition of $V^+$, there is a probability $P^+$ of encountering a knowledge $v^+\in V^+$ that is superior to $v^\tau$.
Assuming the samples in $\boldsymbol{v}$ are independent and identically distributed, the probability of having at least one superior knowledge $v^+$ in $\boldsymbol{v}$ is simply $P^+_k=1-\left(1-P^+\right)^k$.
With a growing number of source tasks, we have
\begin{equation}
\lim_{k\to\infty}P^+_k=\lim_{k\to\infty}1-\left(1-P^+\right)^k=1.
\label{eq:probability_useful_inf}
\end{equation}

Given the probabilistically guaranteed existence of $v^+$ in $\boldsymbol{v}$ under the big-source condition, the retrieval function will consistently capture this superior knowledge to achieve a positive performance gain.
Finally, let us examine the mapping function $f_{\mathrm{m}}$, which can lift its infimum of performance gain through a bigger $\delta_u$.
More specifically, an improvement $\delta_u>u^\tau-u_\mathrm{min}$ will raise the infimum to a positive value, provided that an appropriate mapping can be established.
In essence, the appropriateness here hinges on the realizability and optimality discussed in Section \ref{akt:how}.

\end{IEEEproof}

The implication of Theorem 5 is intuitively clear: to achieve positive transfer, one can either actively expand the pool of source tasks to ensure the availability of useful knowledge, or aptly adapt the knowledge given until it becomes advantageous for the target task.
The notion of actively expanding the source pool emerges naturally during the problem formulation process; however, it has yet to be formalized as a methodology for rigorous investigation.
Nevertheless, this implication underpins many works on multiform optimization, which actively create auxiliary tasks as relevant sources tailored to the target task.
In contrast, the idea of knowledge adaptation has been extensively explored in the literature~\cite{lin2023ensemble}.

\subsection{Compliance with the No Free Lunch Theorem}

On the basis of `no free lunch for transfer' in~\cite{scott2023first}, we offer a novel insight that interprets the effectiveness of ETO and traditional evolutionary optimization (EO) from a unified perspective: aligning search biases with the problems of interest is essential for an optimizer to be effective~\cite{wolpert1997no}, as given by

\begin{equation}
\begin{split}
P(v^\tau\mid v^{1:\tau-1},\beta)&=\sum_{g\in\mathcal{G}}P(v^\tau\mid v^{1:\tau-1},\beta, g)P(g),\\
&=\vec{\boldsymbol{a}}_{v^{1:\tau-1},\beta}\cdot\vec{ \boldsymbol{p}},
\end{split}
\label{eq:nfl_performance}
\end{equation}
where $g\in\mathcal{G}$ is the objective function of an elementary task $g$ in the optimization task family $\mathcal{G}$ of interest, $P(v^\tau\mid v^{1:\tau-1},\beta,g)$ denotes the performance of an optimizer with specific search biases $\beta$ iterated $\tau-1$ times on the function $g$, which is the conditional
probability of obtaining a particular knowledge candidate $v^\tau$ under the stated conditions, $P(g)$ denotes the probability distribution function of $g$ in $\mathcal{G}$, $P(v^\tau\mid v^{1:\tau-1},\beta)$ is the performance on $\mathcal{G}$, $\vec{\boldsymbol{a}}_{v^{1:\tau-1},\beta}(g)\equiv P(v^\tau\mid v^{1:\tau-1},\beta, g)$ is the $\mathcal{F}$-space vector of performance, and $\vec{\boldsymbol{p}}(f)\equiv P(f)$ denotes the $f$ components of $\vec{\boldsymbol{a}}_{v^{1:\tau-1},\beta}(g)$.

\begin{figure}[htbp]
	\centerline{\includegraphics[width=3.3in]{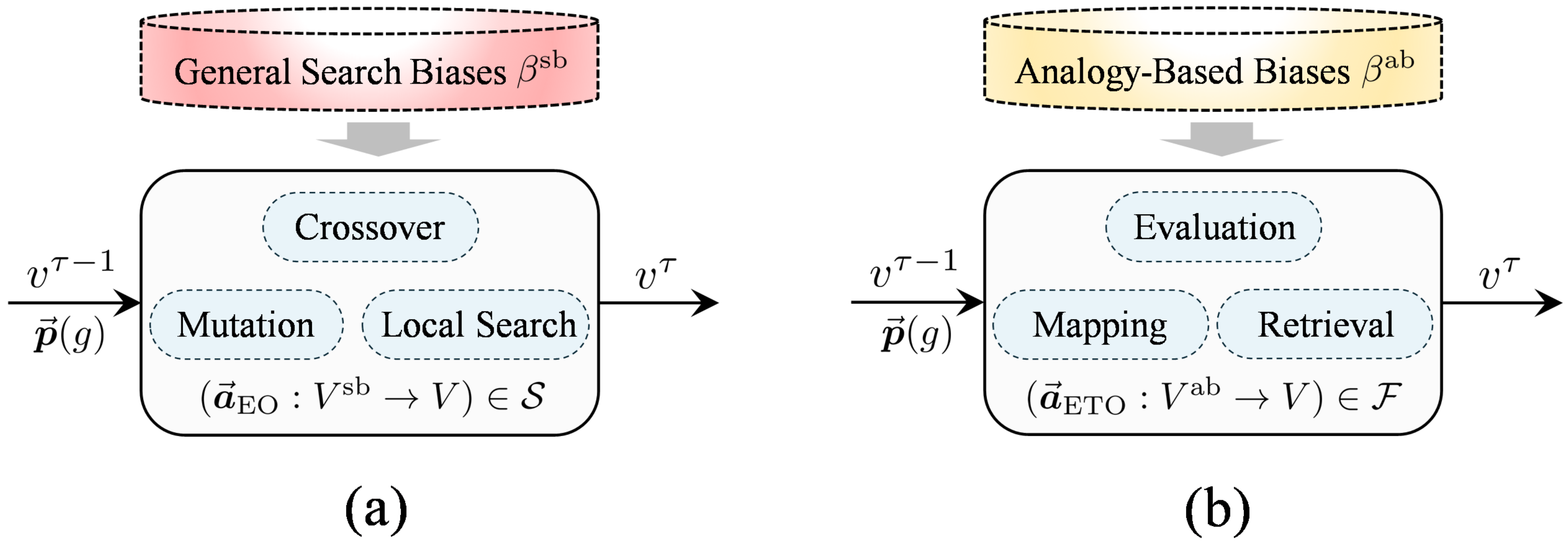}}
	\caption{An illustrative comparison of how EO and ETO incorporate their biases to achieve effective problem-solving: (a) EO; (b) ETO.}
	\label{fig:nfl}
\end{figure}

From a geometric perspective, Eq. \eqref{eq:nfl_performance} shows that the performance of an optimizer $\vec{\boldsymbol{a}}$ with specific search biases $\beta$ is determined by the magnitude of its projection onto $\vec{\boldsymbol{p}}$, i.e., by how aligned the algorithm $\vec{\boldsymbol{a}}$ is with the task family $\vec{\boldsymbol{p}}$ of interest\footnote{A more detailed discussion in this regard can be found in~\cite{wolpert1997no}.}.
This enables us to interpret the effectiveness of EO and ETO from a unified perspective by exploring their distinct approaches to incorporating search biases, as compared in Fig. \ref{fig:nfl}.
In particular, as shown in Fig. \ref{fig:nfl}(a), the search biases used in EO are primarily reflected in the configuration of parent individuals as input and evolutionary operators (e.g., crossover) as intermediate processing, as given by $(\vec{\boldsymbol{a}}_{\mathrm{EO}}:V^{\mathrm{sb}}\to V)\in\mathcal{S}$, where $\vec{\boldsymbol{a}}_{\mathrm{EO}}\in\mathcal{S}$ denotes an evolutionary optimizer with certain operators that generate $v^\tau$ based on the specified input in $V^{\mathrm{sb}}$, $\mathcal{S}$ denotes the set of all valid evolutionary optimizers.
Various configurations in this context are referred to as \emph{general search biases} $\beta^{\mathrm{sb}}$.
Following this perspective, we can regard analogy-based ETO as a specialized form of EO, where search biases are integrated through the configuration of source-target data and a knowledge transfer method $f\in\mathcal{F}$ serving as the intermediate processing stage, as shown in Fig. \ref{fig:nfl}(b) and formally given by $(\vec{\boldsymbol{a}}_{\mathrm{ETO}}:V^{\mathrm{ab}}\to V)\in\mathcal{F}$, where $\vec{\boldsymbol{a}}_{\mathrm{ETO}}$ is an analogy-based ETO optimizer that uses $f\in\mathcal{F}$ to generate $v^\tau$ based on the designated input in $V^{\mathrm{ab}}$.
Various configurations related to $f\in\mathcal{F}$ and its input are referred to as \emph{analogy-based biases} $\beta^{\mathrm{ab}}$.
Although $\beta^{\mathrm{sb}}$ and $\beta^{\mathrm{ab}}$ differ in their manifestations, both are typically derived from our understanding of $g\in\mathcal{G}$ being addressed.
Intuitively, a deeper understanding in this context will enable the algorithm $\vec{\boldsymbol{a}}_{\beta}$ to align more closely with $\vec{\boldsymbol{p}}$, resulting in improved overall performance on $\mathcal{G}$, as indicated by Eq. \eqref{eq:nfl_performance}.

Our insights here not only demonstrate the compliance of analogy-based knowledge transfer with the no free lunch theorem for optimization but also highlight the conditional superiority of ETO over EO.
Specifically, the performance gain of analogy-based knowledge transfer on a problem of interest will be positive if and only if the inductive biases of analogical reasoning align more effectively with the problem than the general search biases in traditional EO, thereby facilitating more effective problem-solving through positive transfer.
This well explains why a specific knowledge transfer method may improve an EO solver $\vec{\boldsymbol{a}}_{A}$, but it could negatively impact the search optimization of another solver $\vec{\boldsymbol{a}}_{B}$ when $\beta^{\mathrm{sb}}_B$ is superior to $\beta^{\mathrm{sb}}_A$.
In such cases, it is more challenging for the knowledge transfer method to catch up with the faster search speed of $\vec{\boldsymbol{a}}_{B}$ compared to that of $\vec{\boldsymbol{a}}_{A}$.
This phenomenon has also been theoretically demonstrated in Section IV-C of~\cite{xue2024surrogate}.


\section{Conclusions}
\label{sec:con}

In this study, we have laid the groundwork for a deeper understanding of analogy-based ETO by addressing the theoretical gaps that have emerged alongside the rapid development of new algorithms.
Our exploration began with an introduction to analogical reasoning, connecting its subprocesses to the key issues in ETO.
Then, we have established the theoretical foundations of analogy-based knowledge transfer by separately examining the principles underlying the three subprocesses.

With the established theoretical foundations, we have come up with two important theorems concerning the performance gain of analogy-based knowledge transfer.
We anticipate that these theorems will facilitate the design of relevant analogy-based ETO algorithms aimed at achieving interpretable and generalizable performance.
Moreover, we have offered a novel insight into analogy-based ETO, demonstrating that its superiority over traditional EO arises from analogy-based biases that align more closely with the problem of interest, compared to the general search biases inherent in EO.
Therefore, based on the analogy-based principles discussed in Section \ref{sec:akt}, it is beneficial to conduct empirical studies that align analogy-based biases more closely with the problem at hand.
Thus, building on the analogy-based principles outlined in Section \ref{sec:akt}, conducting relevant empirical studies can help better align analogy-based biases with the problem of interest~\cite{xue2023solution,scott2024varying}.
This alignment facilitates more effective problem-solving through ETO, akin to incorporating the inductive bias of sparsity into learning tasks~\cite{deng2018motor}.

In future work, we will explore principled methods for automatically discovering analogy-based biases relevant to our problems of interest, facilitating trustworthy ETO that consistently outperforms EO.
Additionally, we are interested in exploring the theoretical foundations of ETO based on other methods of reasoning, such as inductive generalization, prediction and causal inference.




\footnotesize
\bibliography{mybib}

\begin{thebibliography}{10}
\providecommand{\url}[1]{#1}
\csname url@samestyle\endcsname
\providecommand{\newblock}{\relax}
\providecommand{\bibinfo}[2]{#2}
\providecommand{\BIBentrySTDinterwordspacing}{\spaceskip=0pt\relax}
\providecommand{\BIBentryALTinterwordstretchfactor}{4}
\providecommand{\BIBentryALTinterwordspacing}{\spaceskip=\fontdimen2\font plus
\BIBentryALTinterwordstretchfactor\fontdimen3\font minus
  \fontdimen4\font\relax}
\providecommand{\BIBforeignlanguage}[2]{{%
\expandafter\ifx\csname l@#1\endcsname\relax
\typeout{** WARNING: IEEEtran.bst: No hyphenation pattern has been}%
\typeout{** loaded for the language `#1'. Using the pattern for}%
\typeout{** the default language instead.}%
\else
\language=\csname l@#1\endcsname
\fi
#2}}
\providecommand{\BIBdecl}{\relax}
\BIBdecl

\bibitem{tan2021evolutionary}
K.~C. Tan, L.~Feng, and M.~Jiang, ``Evolutionary transfer optimization-a new
  frontier in evolutionary computation research,'' \emph{IEEE Computational
  Intelligence Magazine}, vol.~16, no.~1, pp. 22--33, 2021.

\bibitem{gupta2017insights}
A.~Gupta, Y.-S. Ong, and L.~Feng, ``Insights on transfer optimization: Because
  experience is the best teacher,'' \emph{IEEE Transactions on Emerging Topics
  in Computational Intelligence}, vol.~2, no.~1, pp. 51--64, 2018.

\bibitem{feng2017autoencoding}
L.~Feng, Y.-S. Ong, S.~Jiang, and A.~Gupta, ``Autoencoding evolutionary search
  with learning across heterogeneous problems,'' \emph{IEEE Transactions on
  Evolutionary Computation}, vol.~21, no.~5, pp. 760--772, 2017.

\bibitem{xue2023solution}
X.~Xue, C.~Yang, L.~Feng, K.~Zhang, L.~Song, and K.~C. Tan, ``Solution transfer
  in evolutionary optimization: An empirical study on sequential transfer,''
  \emph{IEEE Transactions on Evolutionary Computation}, vol.~28, no.~6, pp.
  1776--1793, 2023.

\bibitem{gupta2015multifactorial}
A.~Gupta, Y.-S. Ong, and L.~Feng, ``Multifactorial evolution: Toward
  evolutionary multitasking,'' \emph{IEEE Transactions on Evolutionary
  Computation}, vol.~20, no.~3, pp. 343--357, 2015.

\bibitem{feng2018evolutionary}
L.~Feng, L.~Zhou, J.~Zhong, A.~Gupta, Y.-S. Ong, K.-C. Tan, and A.~K. Qin,
  ``Evolutionary multitasking via explicit autoencoding,'' \emph{IEEE
  Transactions on Cybernetics}, vol.~49, no.~9, pp. 3457--3470, 2018.

\bibitem{feng2024review}
Y.~Feng, L.~Feng, X.~Xue, S.~Kwong, and K.~C. Tan, ``A review on evolutionary
  multiform transfer optimization,'' in \emph{2024 IEEE Congress on
  Evolutionary Computation (CEC)}.\hskip 1em plus 0.5em minus 0.4em\relax IEEE,
  2024, pp. 1--8.

\bibitem{jiang2017transfer}
M.~Jiang, Z.~Huang, L.~Qiu, W.~Huang, and G.~G. Yen, ``Transfer learning-based
  dynamic multiobjective optimization algorithms,'' \emph{IEEE Transactions on
  Evolutionary Computation}, vol.~22, no.~4, pp. 501--514, 2017.

\bibitem{liu2022evolutionary}
S.~Liu, Q.~Lin, L.~Feng, K.-C. Wong, and K.~C. Tan, ``Evolutionary multitasking
  for large-scale multiobjective optimization,'' \emph{IEEE Transactions on
  Evolutionary Computation}, vol.~27, no.~4, pp. 863--877, 2022.

\bibitem{li2024multiobjective}
Y.~Li and W.~Gong, ``Multiobjective multitask optimization with multiple
  knowledge types and transfer adaptation,'' \emph{IEEE Transactions on
  Evolutionary Computation}, vol.~29, no.~1, pp. 205--216, 2024.

\bibitem{gao2023distributed}
K.~Gao, C.~Yang, J.~Ding, K.~C. Tan, and T.~Chai, ``Distributed knowledge
  transfer for evolutionary multitask multimodal optimization,'' \emph{IEEE
  Transactions on Evolutionary Computation}, vol.~28, no.~4, pp. 1141--1155,
  2023.

\bibitem{kawakami2024evolutionary}
S.~Kawakami, K.~Takadama, and H.~Sato, ``Evolutionary constrained
  multi-factorial optimization based on task similarity,'' in \emph{2024 IEEE
  Congress on Evolutionary Computation (CEC)}.\hskip 1em plus 0.5em minus
  0.4em\relax IEEE, 2024, pp. 1--8.

\bibitem{ji2025similar}
J.~Ji, X.~Zhang, C.~Yang, X.~Li, and G.~Sui, ``A similar environment transfer
  strategy for dynamic multiobjective optimization,'' \emph{Information
  Sciences}, p. 122018, 2025.

\bibitem{liaw2017evolutionary}
R.-T. Liaw and C.-K. Ting, ``Evolutionary many-tasking based on biocoenosis
  through symbiosis: A framework and benchmark problems,'' in \emph{2017 IEEE
  Congress on Evolutionary Computation (CEC)}.\hskip 1em plus 0.5em minus
  0.4em\relax IEEE, 2017, pp. 2266--2273.

\bibitem{feng2023multi}
Y.~Feng, L.~Feng, S.~Kwong, and K.~C. Tan, ``A multi-form evolutionary search
  paradigm for bi-level multi-objective optimization,'' \emph{IEEE Transactions
  on Evolutionary Computation}, vol.~28, no.~6, pp. 1719--1732, 2023.

\bibitem{friess2021predicting}
S.~Friess, P.~Ti{\v{n}}o, S.~Menzel, B.~Sendhoff, and X.~Yao, ``Predicting
  {CMA-ES} operators as inductive biases for shape optimization problems,'' in
  \emph{2021 IEEE Symposium Series on Computational Intelligence (SSCI)}.\hskip
  1em plus 0.5em minus 0.4em\relax IEEE, 2021, pp. 01--07.

\bibitem{wu2024learning}
S.-H. Wu, Y.~Huang, X.~Wu, L.~Feng, Z.-H. Zhan, and K.~C. Tan, ``Learning to
  transfer for evolutionary multitasking,'' \emph{arXiv preprint
  arXiv:2406.14359}, 2024.

\bibitem{wang2025learning}
C.~Wang, L.~Jiao, J.~Zhao, L.~Li, F.~Liu, and S.~Yang, ``Learning evolution via
  optimization knowledge adaptation,'' \emph{arXiv preprint arXiv:2501.02200},
  2025.

\bibitem{xue2023scalable}
X.~Xue, C.~Yang, L.~Feng, K.~Zhang, L.~Song, and K.~C. Tan, ``A scalable test
  problem generator for sequential transfer optimization,'' \emph{IEEE
  Transactions on Cybernetics}, 2025, accepted, DOI: 10.1109/TCYB.2025.3547565.

\bibitem{hou2024bridging}
Y.~Hou, W.~Ma, A.~Gupta, K.~K. Bali, H.~Ge, Q.~Zhang, C.~A.~C. Coello, and
  Y.-S. Ong, ``Bridging the gap between theory and practice: Benchmarking
  transfer evolutionary optimization,'' \emph{arXiv preprint arXiv:2404.13377},
  2024.

\bibitem{gupta2016landscape}
A.~Gupta, Y.-S. Ong, B.~Da, L.~Feng, and S.~D. Handoko, ``Landscape synergy in
  evolutionary multitasking,'' in \emph{2016 IEEE Congress on Evolutionary
  Computation (CEC)}.\hskip 1em plus 0.5em minus 0.4em\relax IEEE, 2016, pp.
  3076--3083.

\bibitem{lin2024multiobjective}
Q.~Lin, Q.~Wang, B.~Chen, Y.~Ye, L.~Ma, and K.~C. Tan, ``Multiobjective
  many-tasking evolutionary optimization using diversified {Gaussian}-based
  knowledge transfer,'' \emph{IEEE Transactions on Evolutionary Computation},
  2024, accepted, DOI: 10.1109/TEVC.2024.3467048.

\bibitem{cai2021evolutionary}
Y.~Cai, D.~Peng, P.~Liu, and J.-M. Guo, ``Evolutionary multi-task optimization
  with hybrid knowledge transfer strategy,'' \emph{Information Sciences}, vol.
  580, pp. 874--896, 2021.

\bibitem{bali2017linearized}
K.~K. Bali, A.~Gupta, L.~Feng, Y.~S. Ong, and T.~P. Siew, ``Linearized domain
  adaptation in evolutionary multitasking,'' in \emph{2017 IEEE Congress on
  Evolutionary Computation (CEC)}.\hskip 1em plus 0.5em minus 0.4em\relax IEEE,
  2017, pp. 1295--1302.

\bibitem{scott2023first}
E.~O. Scott and K.~A. De~Jong, ``First complexity results for evolutionary
  knowledge transfer,'' in \emph{Proceedings of the 17th ACM/SIGEVO Conference
  on Foundations of Genetic Algorithms}, 2023, pp. 140--151.

\bibitem{wolpert1997no}
D.~H. Wolpert and W.~G. Macready, ``No free lunch theorems for optimization,''
  \emph{IEEE Transactions on Evolutionary Computation}, vol.~1, no.~1, pp.
  67--82, 1997.

\bibitem{bali2019multifactorial}
K.~K. Bali, Y.-S. Ong, A.~Gupta, and P.~S. Tan, ``Multifactorial evolutionary
  algorithm with online transfer parameter estimation: {MFEA-II},'' \emph{IEEE
  Transactions on Evolutionary Computation}, vol.~24, no.~1, pp. 69--83, 2019.

\bibitem{xue2024surrogate}
X.~Xue, Y.~Hu, L.~Feng, K.~Zhang, L.~Song, and K.~C. Tan, ``Surrogate-assisted
  search with competitive knowledge transfer for expensive optimization,''
  \emph{IEEE Transactions on Evolutionary Computation}, 2024, accepted, DOI:
  10.1109/TEVC.2024.3478732.

\bibitem{min2020generalizing}
A.~T.~W. Min, A.~Gupta, and Y.-S. Ong, ``Generalizing transfer {Bayesian}
  optimization to source-target heterogeneity,'' \emph{IEEE Transactions on
  Automation Science and Engineering}, vol.~18, no.~4, pp. 1754--1765, 2020.

\bibitem{liu2024extremo}
J.~Liu, A.~Gupta, C.~Ooi, and Y.-S. Ong, ``{ExTrEMO}: Transfer evolutionary
  multiobjective optimization with proof of faster convergence,'' \emph{IEEE
  Transactions on Evolutionary Computation}, vol.~29, no.~1, pp. 102--116,
  2024.

\bibitem{min2017multiproblem}
A.~T.~W. Min, Y.-S. Ong, A.~Gupta, and C.-K. Goh, ``Multiproblem surrogates:
  Transfer evolutionary multiobjective optimization of computationally
  expensive problems,'' \emph{IEEE Transactions on Evolutionary Computation},
  vol.~23, no.~1, pp. 15--28, 2017.

\bibitem{da2018curbing}
B.~Da, A.~Gupta, and Y.-S. Ong, ``Curbing negative influences online for
  seamless transfer evolutionary optimization,'' \emph{IEEE transactions on
  cybernetics}, vol.~49, no.~12, pp. 4365--4378, 2018.

\bibitem{cao2024competitive}
C.~Cao, X.~Xue, K.~Zhang, L.~Song, L.~Zhang, X.~Yan, Y.~Yang, J.~Yao, W.~Zhou,
  and C.~Liu, ``Competitive knowledge transfer--enhanced surrogate-assisted
  search for production optimization,'' \emph{SPE Journal}, vol.~29, no.~6, pp.
  3277--3292, 2024.

\bibitem{scott2024varying}
E.~O. Scott and K.~A. De~Jong, ``Varying difficulty of knowledge reuse in
  benchmarks for evolutionary knowledge transfer,'' in \emph{2024 IEEE Congress
  on Evolutionary Computation (CEC)}.\hskip 1em plus 0.5em minus 0.4em\relax
  IEEE, 2024, pp. 01--08.

\bibitem{villar2025transfer}
E.~Villar-Rodriguez, E.~Osaba, I.~Oregi, S.~V. Romero, and
  J.~Ferreiro-V{\'e}lez, ``On the transfer of knowledge in quantum
  algorithms,'' \emph{arXiv preprint arXiv:2501.14120}, 2025.

\bibitem{gentner2012analogical}
D.~Gentner and L.~Smith, ``Analogical reasoning,'' in \emph{V. S. Ramachandran
  (Ed.) Encyclopedia of Human Behavior (2nd Ed.)}.\hskip 1em plus 0.5em minus
  0.4em\relax Oxford, UK: Elsevier, 2012, pp. 130--136.

\bibitem{chen2019adaptive}
Y.~Chen, J.~Zhong, L.~Feng, and J.~Zhang, ``An adaptive archive-based
  evolutionary framework for many-task optimization,'' \emph{IEEE Transactions
  on Emerging Topics in Computational Intelligence}, vol.~4, no.~3, pp.
  369--384, 2019.

\bibitem{cao2024global}
C.~Cao, K.~Zhang, X.~Xue, K.~C. Tan, J.~Wang, L.~Zhang, P.~Liu, and X.~Yan,
  ``Global and local search experience-based evolutionary sequential transfer
  optimization,'' \emph{IEEE Transactions on Evolutionary Computation}, 2024,
  accepted, DOI: 10.1109/TEVC.2024.3417325.

\bibitem{xue2020affine}
X.~Xue, K.~Zhang, K.~C. Tan, L.~Feng, J.~Wang, G.~Chen, X.~Zhao, L.~Zhang, and
  J.~Yao, ``Affine transformation-enhanced multifactorial optimization for
  heterogeneous problems,'' \emph{IEEE Transactions on Cybernetics}, vol.~52,
  no.~7, pp. 6217--6231, 2020.

\bibitem{friess2020improving}
S.~Friess, P.~Ti{\v{n}}o, S.~Menzel, B.~Sendhoff, and X.~Yao, ``Improving
  sampling in evolution strategies through mixture-based distributions built
  from past problem instances,'' in \emph{Parallel Problem Solving from
  Nature--PPSN XVI: 16th International Conference, PPSN 2020, Leiden, The
  Netherlands, September 5-9, 2020, Proceedings, Part I 16}.\hskip 1em plus
  0.5em minus 0.4em\relax Springer, 2020, pp. 583--596.

\bibitem{lin2023ensemble}
W.~Lin, Q.~Lin, L.~Feng, and K.~C. Tan, ``Ensemble of domain adaptation-based
  knowledge transfer for evolutionary multitasking,'' \emph{IEEE Transactions
  on Evolutionary Computation}, vol.~28, no.~2, pp. 388--402, 2023.

\bibitem{deng2018motor}
X.~Deng, D.~Li, J.~Mi, F.~Gao, Q.~Chen, J.~Wang, and R.~Liu, ``Motor imagery
  ecog signal classification using sparse representation with elastic net
  constraint,'' in \emph{2018 IEEE 7th Data Driven Control and Learning Systems
  Conference (DDCLS)}.\hskip 1em plus 0.5em minus 0.4em\relax IEEE, 2018, pp.
  44--49.

\end{thebibliography}

\end{document}